\documentclass[runningheads]{llncs}
\usepackage{graphicx}
\usepackage{booktabs}
\usepackage{multirow}
\usepackage[normalem]{ulem}
\usepackage[sectionbib, numbers]{natbib}
\useunder{\uline}{\ul}{}

\usepackage{amsmath}
\usepackage{framed}
\usepackage{mdframed}
\usepackage{lipsum}
\usepackage{hyperref}

\usepackage{footnote}
\usepackage{enumitem}
\usepackage{longtable}
\usepackage{color}
\usepackage[table,dvipsnames]{xcolor}
\usepackage{adjustbox}
\usepackage{marvosym}

\usepackage{diagbox}
\newcolumntype{M}[1]{>{\centering\arraybackslash}m{#1}}
\newcommand*\rot{\rotatebox{90}}
\definecolor{effort}{HTML}{6DFCFF}
\definecolor{outcome}{HTML}{F4CCCC}
\begin{document}

\title{Using Large Language Models to Provide Explanatory Feedback to Human Tutors}

\titlerunning{Automatic Explanatory Feedback to Human Tutors}

%
%
\author{Jionghao Lin \inst{1, 2} \and
Danielle R. Thomas \inst{1}\and
Feifei Han  \inst{3}\and
Shivang Gupta \inst{1} \and
Wei Tan \inst{2} \and
Ngoc Dang Nguyen \inst{2} \and
Kenneth R. Koedinger \inst{1}}
\authorrunning{Lin et al.}
%
\institute{Carnegie Mellon University, Pittsburgh, PA, USA\\
\email{\{Jionghao,Drthomas,Shivang,Koedinger\}@cmu.edu}\\ \and
Monash University, Clayton, VIC, Australia \\
\email{\{wei.tan2,dan.nguyen2\}@monash.edu}\\ \and
University of Toronto, Toronto, Ontario, Canada\\
\email{feifei.han@mail.utoronto.ca}
}
\maketitle              
\begin{abstract}
Research demonstrates learners engaging in the process of producing explanations to support their reasoning, can have a positive impact on learning. However, providing learners real-time explanatory feedback often presents challenges related to classification accuracy, particularly in domain-specific environments, containing situationally complex and nuanced responses. We present two approaches for supplying tutors real-time feedback within an online lesson on how to give students effective praise. This work-in-progress demonstrates considerable accuracy in binary classification for corrective feedback of effective, or effort-based ($F_1 score$ = 0.811), and  ineffective, or outcome-based ($F_1 score$ = 0.350), praise responses. More notably, we introduce progress towards an enhanced approach of providing explanatory feedback using large language model-facilitated named entity recognition, which can provide tutors feedback, not only while engaging in lessons, but can potentially suggest real-time tutor moves. Future work involves leveraging large language models for data augmentation to improve accuracy, while also developing an explanatory feedback interface. 


\end{abstract}
\section{Introduction}

Tutoring is among the most highly adaptable and consistently successful interventions to increase student learning \cite{kraft2021blueprint, nickow2020impressive}. However, despite the known positive impacts of tutoring on achievement, there is a lack of qualified and skilled tutors outside of private, high-income communities, ready to provide content and socio-motivational support to students \cite{kraft2021blueprint}. Due to the shortage of professional tutors, often certified teachers and paraprofessionals, the focus has shifted to preparing novice tutors, such as community volunteers, retired adults, and college students \cite{nickow2020impressive}. The demand for professional development personalized to meet the needs of  nonprofessional and novice  tutors is high \cite{nickow2020impressive}, with training on social-emotional learning, relationship building, and attending to student motivation and self-efficacy as common topics requested among unskilled tutors \cite{thomas2023tutor}. Online, scenario-based lessons on these topics have been developed to provide situational experiences to inexperienced tutors \cite{thomas2023tutor} and pre-service teachers \cite{thompson2019teacher}. The ability to administer real-time explanatory feedback within constructed-response questions dealing with common tutoring scenarios (e.g., a student struggling with motivation) is powerful. Immediate feedback on errors, similar to the feedback received while engaging in the deliberate practice of responding to situational judgment tests, is described as a ``favorable learning condition,'' supporting learning \cite[p.~5]{koedinger2023astonishing}.

\begin{figure}[b]
\centering
\vspace{-9mm}
\includegraphics[width=0.9\textwidth]{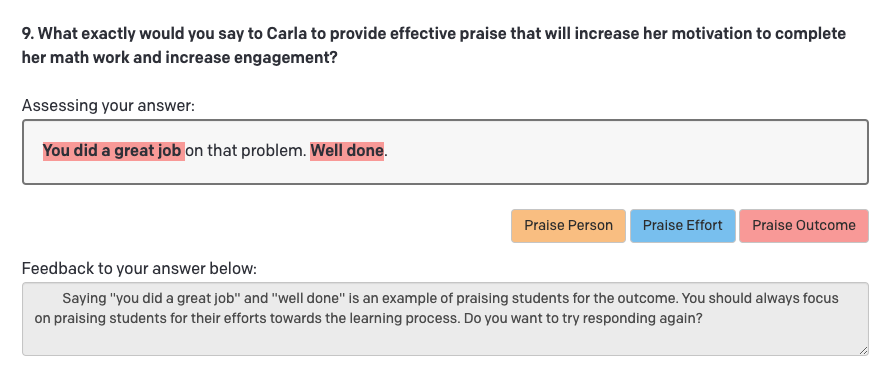}
\caption{An example of providing a tutor templated feedback that does not contain the desire responses of praising students for their efforts.} \label{fig1}
\vspace{-3mm}
\end{figure}
We present a method of providing tutors real-time explanatory feedback harnessing large language models (LLMs). Our approach employs a template-based strategy leveraging named entity recognition (NER), a subtask of natural language processing, that classifies similar pieces of information \cite{li2020survey}. By tagging similar pieces of information, called named entities (NEs), for effective and ineffective tutor responses, NER becomes a suitable and viable method for delivering tutors feedback. For example, classifying desired and less-desired tutor responses on how to effectively praise students yields the following NEs: praising for effort, or process-focused praise (\textit{Effort}); ability- or outcome-focused praise (\textit{Outcome}); and person-based praise (\textit{Person}). Using NER, segments of tutor responses can be systematically identified aligning with the appropriate NEs. For instance, in Fig. \ref{fig1} the tutor response \textit{``Good job! You got the right answer, and you stuck with it''} tags tutor utterances to produce the following NEs: \textit{``Good job''} (\textit{Outcome}) and \textit{``stuck with it''} (\textit{Effort}). Tagged NEs can be used to create the corresponding templated feedback: \textit{``Saying [insert Effort] is a nice example of process-focused praise, which praises students for their effort.''} Conversely, templated feedback for a less-desired response could be: \textit{``Saying [insert Outcome] is praising students for the outcome. You should focus on praising the students for their effort and process towards learning. Do you want to try responding again?''} The research-recommended approach, representing the desired tutor response and, commonly observed, less-desired responses can be tagged to corresponding NEs. By tagging pieces of information, aligning with tutoring approaches for responding to the given scenarios, NER becomes a suitable method for generating templated feedback to tutors.

Automatic short answer grading, or the process of automatically scoring learner answers to constructed-responses questions (often applying several machine learning models), has received notable attention due to advances in AI-based technologies \cite{zhang2022automatic}. Most automatic short answer grading methods follow a two-step approach: 1) using a representation, or training set, of learner responses to train the model, using natural language processing methods, and 2) labeling responses via a machine learning classifier to predict the learner's score or performance \cite{zhang2022automatic}. Presently, advanced approaches using LLMs to process learner responses are taking precedence over traditional, human-identified feature analysis \cite{zhang2022automatic}. Despite the advantages to using LLMs, there are several limitations: the model not being well-adapted for the nuanced and varied responses among the learner population; the requirement of having to train one model per question, with related or follow-up questions being treated as mutually exclusive \cite{zhang2022automatic}; and, the need for a large number of tutor responses in the representation dataset. 
This workshop paper presents a method of providing corrective and explanatory feedback to tutors participating in an online lesson on giving students effective praise. This work-in-progress introduces an ongoing effort to enhance approaches towards automatic short answer grading using LLMs, enabling NER for identifying relevant components of a tutor response. The primary research questions addressed, include: 

\begin{itemize}[leftmargin=.42in]
    \item[\textbf{RQ1:}] Can we apply a binary classification method for effectively labeling tutor responses, as effective or ineffective, to provide corrective feedback?
    \item[\textbf{RQ2:}] How can we enhance past approaches of providing explanatory feedback using LLM-facilitated named entity recognition to administer templated feedback identifying relevant parts of the tutor response?  
\end{itemize}

\section{Related Work}
\subsection{Feedback Generation}
Feedback can have a profound impact on learning and achievement; however, its influence may be beneficial or detrimental to learning depending on type and delivery \cite{hattie2007power, ryan2021designing, lin2023learner}. Feedback is most beneficial within the learning context, delivered after the learner has engaged with the initial instruction, and when addressing misconceptions or faulty reasoning \cite{hattie2007power}. Immediate, explanatory feedback, or feedback detailing the reasoning why a response is desired or not, assists learners with participating in deliberate practice. The online lesson has tutors engaging in the deliberate practice of responding to a common tutoring scenario (i.e., a student struggling to stay motivated) by asking them how to best respond. Tutors then explain their reasoning and observe the most-desired approach receiving feedback on their chosen selected response option \cite{chine2022development, thomas2023tutor}. An expansion of this previous work is to provide explanatory feedback to tutors on their textual replies to the constructed-response questions. Generating explanatory feedback to tutors using enhanced approaches, such as using LLM-facilitated NER, shows promise as a method of providing accurate and timely feedback to tutors. The creation of templated feedback, including specific references to desired and less-desired elements of the tutor responses, is influenced by earlier results on the effectiveness of having a rich, data-driven error diagnosis taxonomy driving template-based feedback \cite{aleven2001towards}. 

\subsection{Named Entity Recognition}
A named entity (NE) is a word or phrase distinct from a set of words that have similar attributes \cite{li2020survey}. For example, in the text \textit{``John said that Pittsburgh is wonderful in the winter''},  \textit{``John''}, \textit{``Pittsburgh''}, and \textit{``winter''} are considered NEs, which represent a person, location, and time, respectively. Named entity recognition (NER) is a fundamental task in natural language processing, which aims to automatically locate NE in the text and classify them into different categories such as person, organization, and location \cite{li2020survey}. In the example, to identify the NE \textit{``Pittsburgh''}, a NER model first locates the position of \textit{``Pittsburgh''} in the text and then classifies the entity \textit{``Pittsburgh''} into the category of location. As discussed by \cite{li2020survey}, there are two primary categories of name entities: (i) generic (e.g., person and organization) and (ii) domain-specific (e.g., enzymes and genes). Since the present work aims to investigate the potential of the NER model in providing explanatory feedback based on learning principles in the previous research \cite{thomas2023tutor}, we focus on the domain-specific NER model scheme. In the educational domain, researchers have conducted NER for automatic text assessment \cite{walter2022increasing}. However, the use of the NER model is rarely used in feedback generation. Therefore, our approach aims to employ a NE recognition model to highlight the NEs within tutors' responses, which can be used to create templated explanatory feedback to tutors to increase tutor learning. 

\vspace{-3mm}
\section{Method}
\subsection{Dataset}
The dataset consisted of tutor responses to constructed-response questions within the \textit{Giving Effective Praise} lesson, comprising a total of 65 volunteer tutors. The tutor demographics are: 52\% White, 18\% Asian, 52\% male, and slightly more than half are reportedly 50 years of age or older. The development of \textit{Giving Effective Praise} involved collaboration between the tutoring organization's director and researchers to ensure accurate operationalization of effective praise strategies within the tutoring environment. Lesson scenarios were chosen to enhance face validity by aligning with typical situations encountered by tutors in the field \cite{thomas2023tutor}. \textit{Giving Effective Praise} aims to support tutors with increasing student motivation by providing effective praise, identifying its key features, and employing strategies to deliver praise and feedback. In accordance with the lesson's construct,  \textit{effective} praise should be: (1) sincere, earned, and truthful; (2) specific by giving details of a student's strengths; (3) immediate, with praise given right after the student's action; (4) authentic, avoiding repetitive phrases like \textit{``great job''} which diminishes meaning and becomes predictable, and (5) focused on the learning process rather than innate ability \cite{thomas2023tutor}.

Based on characteristics of praise types in the literature, tutor praise statements can be categorized into three different types: effort-based (\textit{Effort}), outcome-based (\textit{Outcome}), and person-based (\textit{Person}). Effort-based praise is a research-shown productive praise type, focusing on the learning process  (e.g., \textit{``I like how you worked hard to...''}). Outcome-based praise showcases student's achievements, such as getting an A on an assignment or getting a problem correct, and is often, but not always, associated with unproductive praise (e.g., \textit{``Great job!''}). Person-based praise suggests student's success is caused from fixed qualities outside of student's control (e.g., \textit{``You are so talented.''}) and, similar to outcome-based praise, is often associated with unproductive praise \cite{kamins1999person}.

The \textit{Giving Effective Praise} dataset contains 129 tutor responses categorized by praise type. Because only one person-based praise statement (i.e., \textit{``You are very smart''}) was identified from the dataset, person-based praise was not included in the analysis. It should be noted that a tutor's response can include more than one praise type. For example, the statement, \textit{``Great job! I like how you worked hard on completing that task,''} encompasses outcome- and effort-based praise statements. Similarly, a tutor response may not contain any praise types such as, \textit{``Let's work together.''}

\subsection{RQ 1: Binary Classification for Corrective Feedback}
\label{rq1:method}
To accurately identify different types of praise, we aim to conduct multi-label classification. Relying on the praise framework proposed by \cite{thomas2023tutor}, we recruited an educational expert to annotate each type of praise for the tutor's responses in a binary form. The distribution of annotated praise in tutor responses are as follows: 52 responses contained effort-based praise only; 29 responses contained both effort- and outcome-based praise; 26 contained outcome-based praise only; and, the remaining 22 responses lacked any mention of neither effort or outcome-based praise. Then, we trained the annotated responses on classifiers. Inspired by the effectiveness of the BERT model on the educational classification tasks such as tutoring dialogue classification \cite{lin2022good, lin2023robust}, we employed the BERT model to identify each type of praise in the tutors' responses.  To train and evaluate the BERT model, we randomly split the dataset (i.e., annotated tutor responses) into training, validation, and testing set in the ratio of 70\%, 10\%, 20\%, respectively, as suggested by \cite{gholamy201870}. The classification performance of the BERT model was measured by accuracy and F1 score.

\subsection{RQ 2: Named Entity Recognition to Generate Explanatory Feedback}
\label{rq2:method}
In order to generate explanatory feedback to tutors, firstly, the categorization of relevant parts of responses need to be identified through use of NEs. We refer to the annotation scheme by Thomas \textit{et al.} \cite{thomas2023tutor} introduced previously and annotate the NEs representing attributes associated with \textit{Effort} and \textit{Outcome}, for 129 tutor responses. In line with the NER annotation in previous works \cite{li2020survey, nguyen2023auc}, we apply the same BIO-tagging scheme to this present work, that is, \textbf{B} represents the beginning position of the NE in the text, \textbf{I} represents the inside position of the NE in the text, and \textbf{O} represents outside of NE. For example, when annotating praise NEs for a tutor's praise \textit{``You are doing a great job''}, the word \textit{``great''} is identified as the beginning (i.e., \textbf{B\textsubscript{out}}) of the NE \textit{Outcome} and \textit{``job''} is identified inside (i.e., \textbf{I\textsubscript{out}}) of the NE. The remaining text in the response is identified as the outside (i.e., \textbf{O}) of the NE.
After annotating the NEs for each tutor's response, we employed the BERT model to identify the NE from the tutors' responses. The dataset (annotated with NEs) was also divided into training, validation, and testing set in the ratio of 70\%:10\%:20\%, respectively. The statistics of NER annotation data are shown in Table \ref{tab:distribution} which presents \textbf{O} as the major tag in our dataset. Informed by the previous study \cite{nguyen2023auc}, predicting \textbf{O} would not enhance the evaluation score of the NER model, our study also did not take accurate predictions on \textbf{O} when calculating the performance score. To measure the NER model performance, we used the F1 score in line with the recent works on NER task \cite{li2020survey, nguyen2023auc}. 

\begin{table*}[!htb]
\centering
\caption{Distribution of named entities for each dataset by praise types. }
\label{tab:distribution}
\resizebox{0.95\textwidth}{!}{%
\renewcommand{\arraystretch}{1.5}
\begin{tabular}{m{.15\textwidth}M{.18\textwidth}M{.18\textwidth}M{.18\textwidth}M{.18\textwidth}M{.18\textwidth}}
\toprule
\multirow{2}{*}{}   & \multicolumn{5}{c}{\textbf{\% Annotation (B/I/O)}}                                              \\ \cmidrule(l){2-6} 
                    & \textbf{O}    & \textbf{B-Outcome} & \textbf{I-Outcome} & \textbf{B-Effort} & \textbf{I-Effort} \\ \midrule
\textbf{Full}       & 2380 (76.5\%) & 53 (1.7\%)         & 114 (3.7\%)        & 80 (2.6\%)        & 484  (15.6\%)     \\ \midrule
\textbf{Training}   & 1661 (76.5\%) & 38 (1.8\%)         & 75 (3.5\%)         & 58 (2.7\%)        & 338 (15.6\%)      \\ \midrule
\textbf{Validation} & 226 (75.6\%)  & 6 (2.0\%)          & 19 (6.4\%)         & 6 (2.0\%)         & 42 (14.0\%)       \\ \midrule
\textbf{Testing}    & 493 (76.8\%)  & 9 (1.4\%)          & 20 (3.1\%)         & 16 (2.5\%)        & 104 (16.2\%)      \\ \bottomrule
\end{tabular}
}
\vspace{-3mm}
\end{table*}

\setcounter{footnote}{0} 
\section{Results}
\subsection{RQ1: Identifying the correct type of praise}
First, a multi-label classification was implemented by using the case-sensitive BERT base model\footnote{https://huggingface.co/bert-base-cased} \cite{devlin2019bert} to identify the effective type (i.e., \textit{Effort} and \textit{Outcome}) of praise from tutor responses. To minimize the potential impact of random variation, the model was trained on 10 different random seeds and performance was evaluated using the classification of identifying each type of praise. Table \ref{tab:rq1_results} illustrated the effectiveness of the BERT model in accurately tagging \textit{Effort}, demonstrating notably high performance with an average classification accuracy of 0.731 and F1 score of 0.811. The results indicated that the BERT model could effectively tag \textit{Effort}, which could further help the provision of corrective feedback to inform the novice tutors on providing effort-based praise. However, the BERT model's performance in tagging \textit{Outcome} was less successful. The average F1 score for recognizing \textit{Outcome} was 0.350 with a standard deviation of 0.235. As described the distribution of annotated praise in Section \ref{rq1:method}, the number of tutor responses tagging \textit{Outcome} might be inadequate for the BERT model to identify the responses containing \textit{Outcome} accurately. Additionally, the standard deviation of classification performance for tagging \textit{Outcome} (\textit{SD} = 0.235) was four times larger compared to the standard deviation of tagging \textit{Effort} (\textit{SD} = 0.046). The possible explanation for this could be the inadequate number of \textit{Outcome} instances in the test set. Thus, future studies should annotate more tutor responses labeling \textit{Outcome} to improve the BERT model's performance.

\begin{table*}[!htb]
\centering
\caption{Classification performance of BERT model in identifying praise type. The results show the average performance taken from ten random seeds. Standard errors from these experiments are indicated with subscripts.}
\label{tab:rq1_results}
\resizebox{0.7\textwidth}{!}{%
\renewcommand{\arraystretch}{1.5}
\begin{tabular}{M{.2\textwidth}M{.25\textwidth}M{.25\textwidth}}
\toprule
\textbf{Praise type} & \textbf{Accuracy} & \textbf{F1 Score} \\ \midrule
\textit{Effort}      & $0.731_{0.077}$        & $0.811_{0.046}$        \\ \midrule
\textit{Outcome}     & $0.596_{0.089}$        & $0.350_{0.235}$        \\ \bottomrule
\end{tabular}
}
\vspace{-3mm}
\end{table*}

\subsection{RQ 2: Identifying and labeling praise statements in tutor responses}

The BERT model was employed in using the NER approach. To mitigate random variation, 10 different random seeds were used to evaluate performance of the NER model, ensuring reliable estimations of the model's performance. The average F1 score of the model was 0.202 and the standard deviation was 0.039, with the model effectively identifying certain praise entities. Table \ref{tab:ner_result} presented examples of tutor responses with labeled utterances associated with the corresponding NE, displaying \textit{Effort}, (highlighted in \colorbox{effort}{blue}) and \textit{Outcome} (highlighted in \colorbox{outcome}{red}). \textit{Case 1} showed that the model could accurately identify the location of the praise in the text (i.e., the text highlighted in \colorbox{effort}{blue}) and predict the accurate entity type (i.e., \textit{Effort}). Then, the model failed to annotate the NEs for some responses (e.g., \textit{Case 2} in Table \ref{tab:ner_result}). It should be noted that the classification performance of the NER model still had space to improve the performance. One of the major reasons was that the annotated dataset was limited or low-resourced \cite{lin2023robust, nguyen2023auc}. The model might not have a sufficient dataset to train and test the model performance. In Section \ref{future}, we summarized two solutions to enhance the NER model's performance: \textit{i)} data augmentation approaches; \textit{ii)} and AUC Maximization approaches. 
\vspace{-3mm}

\begin{table*}[!htb]
\centering
\caption{Examples of tutor responses from the test dataset, along with named entity prediction. \textbf{True} and \textbf{Pred} stand for the true and predicted named entity, respectively. Notice in \textit{Case 2}, the failure of the predicted model to annotate the statement associated with \textit{Outcome}, possibly attributed to a limited or low-resourced dataset.}
\vspace{-2mm}
\label{tab:ner_result}
\resizebox{1\textwidth}{!}{%
\renewcommand{\arraystretch}{1.7}
\begin{tabular}{m{.05\textwidth}m{.1\textwidth}m{.48\textwidth}M{.2\textwidth}M{.2\textwidth}}
\toprule
\textit{}                        & \textbf{}     & \multirow{1.5}{*}{\textbf{Examples of Tutor Responses}}                                                                                                 & \textbf{\begin{tabular}[c]{@{}c@{}}Named \\ Entity\end{tabular}}                                            & \textbf{\begin{tabular}[c]{@{}c@{}}Prediction \\ Accuracy\end{tabular}}       \\ \hline
\multirow{3}{*}{\rot{\textit{Case 1}}} & \textbf{True} & \textit{\colorbox{effort}{Good job working through this} and trying some different approaches.}                                                         & \colorbox{effort}{\textit{Effort}}                                                   & \multirow{2.5}{*}{Accurate}           \\ \cline{2-4}
                                 & \textbf{Pred} & \textit{\colorbox{effort}{Good job working through this} and trying some different approaches.}                                                        & \colorbox{effort}{\textit{Effort}}                                                   &                                     \\ \hline
\multirow{3}{*}{\rot{\textit{Case 2}}} & \textbf{True} & \textit{Try your best to focus on the next step, you're already \colorbox{outcome}{doing great so far}.}                                             & \colorbox{outcome}{\textit{Outcome}}                                                  & \multirow{2.5}{*}{Inaccurate}         \\ \cline{2-4}
                                 & \textbf{Pred} & \textit{Try your best to focus on the next step, you're already doing great so far.}                                              & \textit{None}                                                     &                                     \\ \hline
\multirow{3.5}{*}{\rot{\textit{Case 3}}} & \textbf{True} & \textit{You did it, you did well, \colorbox{outcome}{you got the} \colorbox{outcome}{right answer} and \colorbox{effort}{you stuck with it}, I'm proud of what you have done. \colorbox{outcome}{Good job}.} & \multirow{-1.5}{*}{\textit{\begin{tabular}[c]{@{}c@{}}\colorbox{outcome}{Outcome}\\ \colorbox{effort}{Effort}\end{tabular}}} & \multirow{3.5}{*}{\begin{tabular}[c]{@{}c@{}}Partially\\ Accurate\end{tabular}} \\ \cline{2-4}
                                 & \textbf{Pred} & \textit{You did it, you \colorbox{outcome}{did well}, you got the right answer and \colorbox{effort}{you stuck with it}, \colorbox{effort}{I'm} \colorbox{effort}{proud of what you have done}. \colorbox{outcome}{Good job}.} & \multirow{-1.5}{*}{\textit{\begin{tabular}[c]{@{}c@{}}\colorbox{outcome}{Outcome}\\ \colorbox{effort}{Effort}\end{tabular}}} &                                     \\ \hline
\multirow{3}{*}{\rot{\textit{Case 4}}} & \textbf{True} & \textit{I am glad you asked for help today. We can do this homework together.}                                                     & None                                                              & \multirow{2.5}{*}{Accurate}           \\ \cline{2-4}
                                 & \textbf{Pred} & \textit{I am glad you asked for help today. We can do this homework together.}                                                     & None                                                              &                                     \\ \bottomrule
\end{tabular}
}
\vspace{-3mm}
\end{table*}

Through evaluating and analyzing the results of NER, we noted that there is a need  for a more nuanced measure that can acknowledge: partial overlap (e.g., \textit{Case 3}); and true negative prediction results (e.g., \textit{Case 4}). In Table \ref{tab:ner_result}, \textit{Case 3}, the model's prediction exhibits a degree of accuracy (i.e., \textit{``Good job''} tagged as \textit{Outcome}, \textit{``you stuck with it''} tagged as \textit{Effort}) but lacks complete correctness (i.e., \textit{``I'm proud of what you have done''} mislabeled as \textit{Effort}). Nevertheless, the former accurate labeling of NEs can still be used for guiding tutors in providing effective praise. Thus, the prediction for \textit{Case 3} is deemed partially accurate. Future research endeavors should focus on developing a measure to calculate partial accuracy, such as computing the intersection over union of the number of tokens in the predicted text and the desired text. Additionally, the NER model could also make true negative predictions where the tutor response did not contain any praise entities and was identified as having none of NEs (i.e., \textit{Case 4} in Table \ref{tab:ner_result}). As discussed in Section \ref{rq2:method}, there lacked accurate predictions on the \textbf{O} tag when calculating the classification performance score. However, it is also important to identify the tutor's responses that only contain the \textbf{O} tag (i.e., none of the NEs) since it could indicate that the tutors might not understand how to deliver correct praise. In Table \ref{tab:ner_result}, the tutor response in \textit{Case 4} did not contain any praise entities and this response was not related to any type of praise. The NER model could successfully identify that the response did not contain any type of praise. Based on the model prediction, feedback can be generated to guide tutors on providing praise that corresponds with the \textit{Effort} and \textit{Outcome} named entities

\vspace{-3mm}

\section{Discussion and Conclusion}

The construction of automatic short answer grading with the capability of providing explanatory feedback is a longstanding task towards delivering timely, specific, and personalized feedback to learners. This study employed large language models to facilitate the provision of corrective and explanatory feedback to tutors, with the main findings summarized in two folds: (1) Large language models (e.g., BERT) have the potential to identify the effort-based praise, which can be used to provide corrective feedback to novice tutors on the appropriate use of effort-based praise to students. (2) Large language models-facilitated named entity recognition (NER) can highlight the key terms associated with praise types from tutors' responses. The highlighted terms can then be integrated into template-based feedback, which can provide real-time explanatory feedback to tutors to enhance tutor learning.

\vspace{-3mm}

\subsection{Implications}

\textbf{Incorporation of a binary classifier can provide automatic corrective feedback.} The developed classifier can be used to determine the correctness of novice tutors in providing different types of praise. The predicted classifier results can be further integrated into the provision of corrective feedback, which is essential in the learning process since corrective feedback can assist the feedback recipients in identifying errors and enhancing understanding \cite{butler2013explanation}. Through the integration of the classifier within the system, we aim to provide automatic corrective feedback to tutors in dispensing various forms of praise. \\
\textbf{Providing automatic templated feedback enhances tutor learning.} To better facilitate the provision of corrective feedback, this study further investigated the potential of NER in identifying the words within the tutors' responses that correspond to the correct types of praise (i.e., \textit{Effort} and \textit{Outcome}). The words identified as correct praise in the tutors' responses can be integrated into a system of providing explanatory feedback. Fig. \ref{fig1}, within the Section of Introduction, illustrates an example of providing a tutor templated explanatory feedback using an integrated NER interface. Referencing the interface, when a tutor composes praise that includes effort- and/or outcome-based praise, the system will label the \textit{Effort} and/or \textit{Outcome} NEs and further provide explanatory feedback to the tutor. As informed by the suggestions of effective feedback \cite{butler2013explanation}, incorporating explanations into feedback can help the tutor better understand the lesson objectives and content. To this end, we believe that integrating the NER model into our system could support the tutor's learning process.

\subsection{Limitations and Future Work}
\label{future}
\textbf{Managing low confidence prediction using the feedback interface.} The confidence level of the model's predictions is a critical aspect to consider in real-world applications. The model confidence level could affect people's belief in the model's accuracy \cite{rechkemmer2022confidence}. Thus, when the model presents low confidence in predicting an instance, it poses a challenge. In such situations, it would be beneficial to design the feedback interface that presents the uncertainties to the learner. For example, when a model's confidence level on a prediction is below a certain threshold, our template-based feedback could provide hedged responses, such as \textit{``Saying ``you are committed'' might be an example of praising effort. Do you want to explain your reasoning?''}. This approach not only helps uphold the credibility of the system but also invites learners to engage critically with the predictions. Future work entails providing hedged feedback responses offering learners the opportunity to explain their reasoning, along with other strategies for effectively managing low confidence predictions in the feedback interface.\\
\textbf{Enhancing the evaluation metrics for NER.} As indicated by \textit{Case 3} and \textit{Case 4} in Table \ref{tab:ner_result}, respectively, the mode's prediction may demonstrate partial correctness, for incidences where the predicted text only partially matches the desired text or, true negative predictions, where the tutor's response is accurately predicted to contain no praise entities. We argue that both partial correctness and the true negative predictions are useful  in providing explanatory feedback and thus, both types of predictions should be credited. However, the traditional NER measure (F1 score) might not fully account for partial correctness and true negative predictions. Therefore, future work should explore the development of measures, such as the degree of dissimilarity between sets via calculation of intersection over union \cite{levandowsky1971distance} to account for these cases, thereby leading to a more comprehensive evaluation of the model's performance.\\
\textbf{Improving NER performance through data augmentation.} By examining the model of NER for identifying the praise entity from the lesson of \textit{Give effective praise}, we found that our annotated NEs might not be sufficient to train the model, which is under a low-resource data scenario \cite{nguyen2023auc}. To address this issue, we aim to collect more real-world data and explore widely-used data augmentation approaches (e.g., oversampling and synonyms replacement) \cite{feng2021survey} and ChatGPT-generated training instances \cite{thomas2023aied} to improve NER model performance. \\
\textbf{Robust models for enhancing the performance of NER.} An alternative solution to low-resource dataset is to employ robust machine learning models. As indicated in the statistic of our dataset (Table \ref{tab:distribution}), more than 70\% of annotations were annotated as the \textbf{O} tag, which was highly imbalanced. To achieve satisfactory performance under the low-resource and imbalance settings, \cite{nguyen2023auc} proposed to use AUC Maximization approaches for the NER task in the biomedical field, which effectively overcome challenges among low-resources and imbalanced class distribution. Thus, we aim to further examine the efficacy of AUC Maximization approaches on recognizing the praise entities.\\
\textbf{Generalizability across tutor lessons.} Our ultimate goal is to provide automatic feedback to all novice tutors who participate in our training sessions and assist them to understand the effective ways to teach students. Thus, a qualified tutor should be able to comprehend all the training lessons. Though our study examined the potentials of labeling tutor responses and providing explanatory feedback for \textit{Giving Effective Praise} lessons, it is necessary to investigate our proposed methods in other lessons such as \textit{Responding to Student's Errors} and \textit{Learning What Students Know} discussed in previous work \cite{thomas2023tutor}.
\vspace{-3mm}
\section{Acknowledgments.}
\vspace{-3mm}
This work is supported with funding from the Richard King Mellon Foundation (Grant \#10851) and the Heinz Endowments (E6291). Any opinions, findings, and conclusions expressed in this material are those of the authors. We would like to thank Dr. Ralph Abboud for his guidance and recommendations regarding the use of large language models and the application of named entity recognition.
\vspace{-3mm}

%
%
%
\bibliographystyle{splncs04}
\bibliography{mybibliography}
%




\end{document}